\documentclass[10pt,twocolumn,letterpaper]{article}

\usepackage{cvpr}
\usepackage{times}
\usepackage{epsfig}
\usepackage{graphicx}
\usepackage{amsmath}
\usepackage{amssymb}

\usepackage{subfigure}
\usepackage{pbox}
\usepackage{wrapfig}
\newcommand{\mycaption}[2]{\caption{\small \textbf{#1.}~#2}}
\usepackage{chngcntr}

\usepackage{booktabs}
\usepackage{array}
\usepackage{xspace}

\usepackage[pagebackref=true,breaklinks=true,letterpaper=true,colorlinks,bookmarks=false]{hyperref}

\cvprfinalcopy 

\def\cvprPaperID{146} 

\ifcvprfinal\fi
\begin{document}

\title{Video Propagation Networks}

\author{
Varun Jampani$^1$, Raghudeep Gadde$^{1, 2}$ and Peter V. Gehler$^{1, 2}$\\
$^1$Max Planck Institute for Intelligent Systems, T{\"u}bingen, Germany\\
$^2$Bernstein Center for Computational Neuroscience, T{\"u}bingen, Germany \\
\texttt{\{varun.jampani,raghudeep.gadde,peter.gehler\}@tuebingen.mpg.de} \\
}

\maketitle

\begin{abstract}
  We propose a technique that propagates information forward
  through video data. The method is conceptually simple and can be applied to tasks that
  require the propagation of structured information, such as semantic labels,
  based on video content.
  We propose a \emph{Video Propagation Network} that processes video frames in an
  adaptive manner. The model is applied online: it propagates information
  forward without the need to access future frames.
  In particular we combine two components, a temporal bilateral network
  for dense and video adaptive filtering, followed by a spatial network to refine
  features and increased flexibility.
  We present experiments on video object segmentation and semantic video segmentation
  and show increased performance comparing to the best previous
  task-specific methods, while having favorable runtime.
  Additionally we demonstrate our approach on an example
  regression task of color propagation in a grayscale video.

\end{abstract}

\vspace{-0.35cm}
\section{Introduction}

\begin{figure}[t]
\begin{center}
\centerline{\includegraphics[width=\columnwidth]{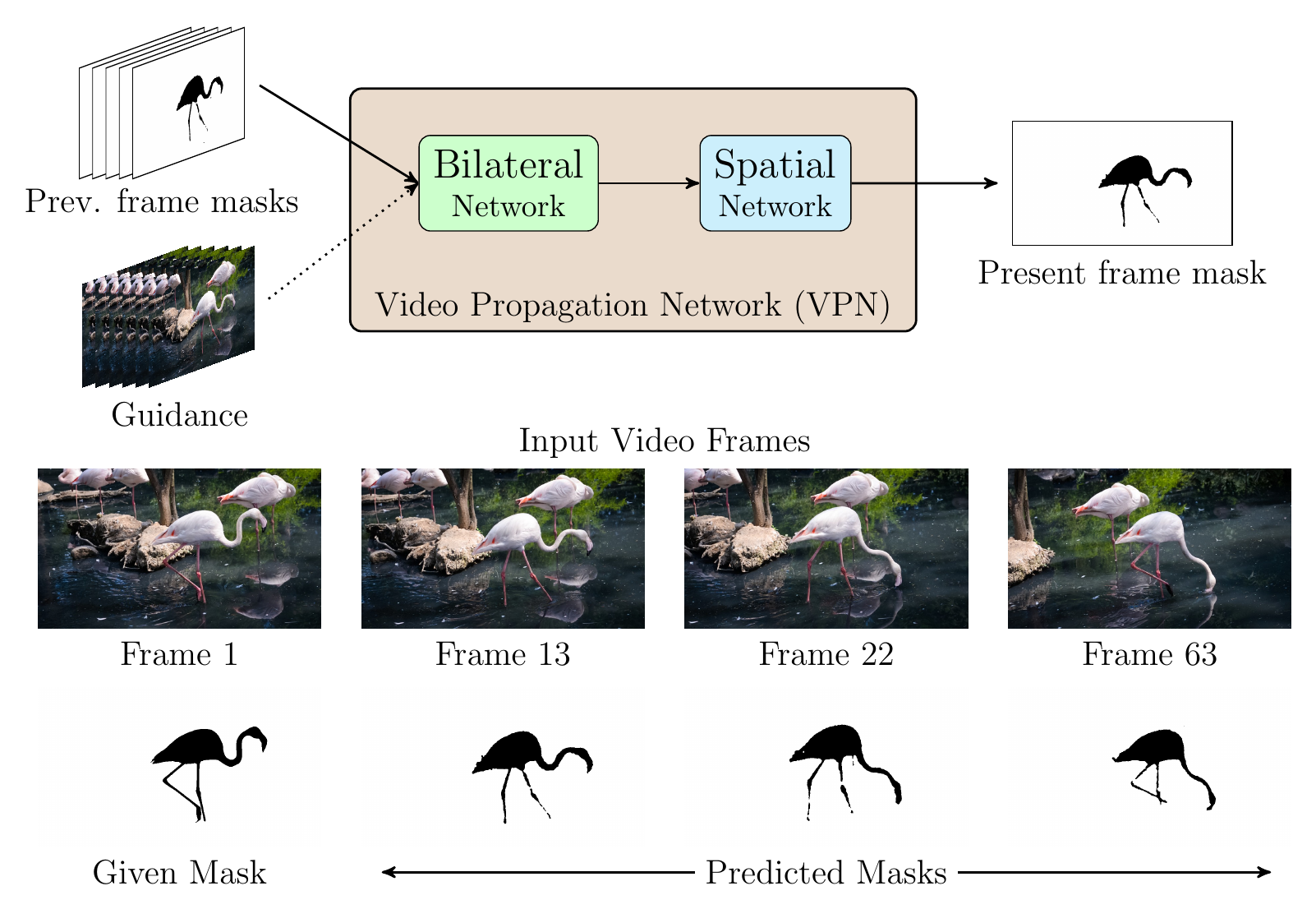}}
  \mycaption{Video Propagation with VPNs} {The end-to-end trained VPN network is composed
  of a bilateral network followed by a standard spatial network and can be used for
  propagating information across frames. Shown here is an example propagation
  of foreground mask from the 1$^{st}$ frame to other video frames.}\label{fig:illustration}
\vspace{-0.5cm}
\end{center}
\vspace{-0.7cm}
\end{figure}

In this work, we focus on the problem of propagating structured information across video frames.~This
problem appears in many forms (e.g., semantic segmentation or depth estimation) and is a pre-requisite for many applications.~An example instance is shown in Fig.~\ref{fig:illustration}.
Given an object mask for the first frame, the problem is to propagate this mask forward
through the entire video sequence.~Propagation of semantic information through time and video
color propagation are other problem instances.

Videos pose both technical and representational challenges.
The presence of scene and camera motion lead to the difficult pixel association problem of optical flow.
Video data is computationally more demanding than static images. A naive per-frame approach would scale at least linear with frames.
These challenges complicate the use of standard convolutional neural networks (CNNs) for video processing.
As a result, many previous works for video propagation use slow optimization based techniques.

We propose a generic neural network architecture that propagates information across
video frames. The main innovation is the use of image adaptive convolutional operations that automatically
adapts to
the video stream content.
~This yields networks that can be applied to several types of information, e.g., labels, colors, etc. and runs online, that is, only requiring current and previous frames.

Our architecture is composed of two components (see Fig.~\ref{fig:illustration}).
A temporal \textit{bilateral network} that performs image-adaptive spatio-temporal dense filtering.
The bilateral network allows to connect densely all pixels from current and previous frames and to propagate associated pixel information to the current frame.
The bilateral network allows the specification of a metric between video pixels and allows a straight-forward integration of temporal information.
This is followed by a standard \textit{spatial CNN} on the bilateral network output to refine and predict for the present video frame.
We call this combination a \textit{Video Propagation Network (VPN)}.
In effect, we are combining video-adaptive filtering with rather small spatial CNNs which
leads to a favorable runtime compared to many previous approaches.

VPNs have the following suitable properties for video processing:

\vspace{-0.5cm}
\paragraph{General applicability:} VPNs can be used to propagate any
type of information content i.e., both discrete
(e.g., semantic labels) and continuous (e.g., color) information across video frames.
\vspace{-0.5cm}
\paragraph{Online propagation:} The method needs no future frames
and can be used for online video analysis.
\vspace{-0.5cm}
\paragraph{Long-range and image adaptive:} VPNs can efficiently handle a large
number of input frames and are adaptive to the video with long-range pixel connections.
\vspace{-0.5cm}
\paragraph{End-to-end trainable:} VPNs can be trained end-to-end, so they
can be used in other deep network architectures.
\vspace{-0.5cm}
\paragraph{Favorable runtime:} VPNs have favorable runtime in comparison to many current
best methods, what makes them amenable for learning with
large datasets.

Empirically we show that VPNs, despite being generic,
perform better than published approaches on video object segmentation
and semantic label propagation while being faster.
VPNs can easily be integrated into sequential per-frame approaches
and require only a small fine-tuning step that can be performed separately.

\section{Related Work}
\label{sec:related}


\paragraph{General propagation techniques}
Techniques for propagating content across image/video pixels are predominantly
optimization based or filtering techniques. Optimization
based techniques typically formulate the propagation as an energy minimization problem
on a graph constructed across video pixels or frames.
A classic example is the color propagation technique from~\cite{levin2004colorization}.
Although efficient closed-form
solutions~\cite{levin2008closed} exists for some scenarios,
optimization tends to be slow due to either large graph structures for videos and/or the use of
complex connectivity.
Fully-connected conditional
random fields (CRFs)~\cite{krahenbuhl2012efficient} open a way for incorporating dense
and long-range pixel connections while retaining fast inference.

Filtering techniques~\cite{kopf2007joint,chang2015propagated,he2010guided} aim to propagate
information with the use of image/video filters resulting in fast runtimes compared
to optimization techniques. Bilateral filtering~\cite{aurich1995non,tomasi1998bilateral} is
one of the popular filters for long-range information propagation.
A popular application is joint bilateral upsampling~\cite{kopf2007joint} that upsamples a low-resolution signal with the use of a high-resolution guidance image.
The works
of~\cite{li2014mean,domke2013learning,kiefel2014permutohedral,jampani_2016_cvpr,zheng2015conditional,schwing2015fully}
showed that one can back-propagate through the bilateral filtering operation for
learning filter parameters~\cite{kiefel2014permutohedral,jampani_2016_cvpr}
or doing optimization in the bilateral space~\cite{barron2015bilateral,barron2015defocus}.
Recently, several works proposed to do upsampling
in images by learning CNNs that mimic edge-aware filtering~\cite{xu2015deep} or
that directly learn to upsample~\cite{li2016deep,hui2016depth}.
Most of these works are confined to images and are either not extendable or computationally
too expensive for videos. We leverage some of these previous works and propose a
scalable yet robust neural network approach for video propagation.
We will discuss more about bilateral filtering,
that forms the core of our approach, in Section~\ref{sec:bilateralfiltering}.

\vspace{-0.5cm}
\paragraph{Video object segmentation}

Prior work on video object segmentation can be broadly categorized into two types:
Semi-supervised methods that require manual annotation to define what is foreground
object and unsupervised methods that does segmentation completely automatically.
Unsupervised techniques such as
~\cite{faktor2014video,li2013video,lee2011key,papazoglou2013fast,wang2015saliency,
zhang2013video,taylor2015causal,dondera2014interactive}
use some prior information about the foreground objects such as
distinctive motion, saliency etc.

In this work, we focus on the semi-supervised task of propagating the foreground
mask from the first frame to the entire video. Existing works
predominantly use graph-based optimization that perform graph-cuts~\cite{boykov2001fast,
boykov2001interactive,shi2000normalized} on video.
Several of these works~\cite{reso2014interactive,
li2005video,price2009livecut,wang2005interactive,kohli2007dynamic,jain2014supervoxel}
aim to reduce the complexity of graph structure with
clustering techniques such as spatio-temporal superpixels and
optical flow~\cite{tsaivideo}.
Another direction was to estimate correspondence between different frame
pixels~\cite{agarwala2004keyframe,bai2009video,lang2012practical} by using
nearest neighbor fields~\cite{fan2015jumpcut} or optical flow~\cite{chuang2002video}.
Closest to our technique are the works of~\cite{perazzi2015fully} and~\cite{marki2016bilateral}.
\cite{perazzi2015fully} proposed to use fully-connected CRF over the
object proposals across frames.~\cite{marki2016bilateral} proposed a
graph-cut in the bilateral space.
Instead of graph-cuts, we learn
propagation filters in the high-dimensional bilateral space.
This results in a more generic architecture and allows integration into other deep networks.
Two contemporary works~\cite{caelles2016one,khoreva2016learning} proposed CNN based
approaches for object segmentation and rely on fine-tuning a deep network
using the first frame annotation of a given test sequence. This could result
in overfitting to the test background.
In contrast, the proposed approach relies only on offline training and thus can be easily adapted
to different problem scenarios as demonstrated in this paper.

\vspace{-0.5cm}
\paragraph{Semantic video segmentation}
Earlier methods such as~\cite{brostow2008segmentation,sturgess2009combining} use structure from motion
on video frames to compute geometrical and/or motion features.
More recent works~\cite{ess2009segmentation,chen2011temporally,de2012line,miksik2013efficient,tripathi2015semantic,
kundu2016feature} construct large graphical models on videos and enforce temporal consistency across frames. \cite{chen2011temporally} used dynamic temporal links in their CRF energy formulation.
\cite{de2012line} proposes to use Perturb-and-MAP random field model with spatial-temporal energy terms
and \cite{miksik2013efficient} propagate predictions across time by learning
a similarity function between pixels of consecutive frames.

In the recent years, there is a big leap in the performance of semantic
segmentation~\cite{long2014fully,chen2014semantic} with the use of CNNs but mostly
applied to images. Recently,~\cite{shelhamer2016clockwork}
proposed to retain the intermediate CNN representations while sliding a image
CNN across the frames. Another approach is to take unary
predictions from CNN and then propagate semantic information across the frames. A recent
prominent approach in this direction is of~\cite{kundu2016feature} which proposes
a technique for optimizing feature spaces for fully-connected CRF.

\vspace{-0.2cm}
\section{Bilateral Filtering}
\label{sec:bilateralfiltering}

We briefly review the bilateral filtering and its extensions that we will need to build VPN.
Bilateral filtering has its roots in image denoising~\cite{aurich1995non,tomasi1998bilateral}
and has been developed as an edge-preserving filter. It has found numerous applications~\cite{paris2009bilateral}
and recently found its way into neural network architectures~\cite{zheng2015conditional,gadde16bilateralinception}.
We will use this filtering at the core of VPN and make use of the image/video-adaptive
connectivity as a way to cope with scenes in motion.

Let $a,\mathbf{a},A$ represent a scalar, vector and matrix respectively.
Bilateral filtering a vectorized image $\mathbf{v} \in \mathbb{R}^n$ having
$n$ image pixels can be viewed as a matrix-vector multiplication
with a filter matrix $W \in \mathbb{R}^{n\times n} $:

\vspace{-0.3cm}
\begin{equation}\label{eq:WtimesV}
  \hat{\mathbf{v}}^i = \sum_{j \in n} W^{i,j} \mathbf{v}^j,
  \vspace{-0.2cm}
\end{equation}

where the filter weights $W^{i,j}$ depend on features $F^i,F^j \in \mathbb{R}^g$ at input pixel indices $i,j$ and $F \in \mathbb{R}^{g \times n}$ for $g$-dimensional features.
For example a Gaussian bilateral filter amounts to a particular choice of $W$ as
$W^{i,j} = \frac{1}{\eta}\exp{(-\frac{1}{2}(F^i-F^j)^{\top}\Sigma^{-1}(F^i-F^j))}$,
where $\eta$ is a normalization constant and $\Sigma$ is covariance matrix.
The choice of features $F$ define the effect of the filter, the way it adapts to image content.
To use only positional features, $F^i=(x,y)^{\top}$, the bilateral filter operation reduces to a spatial Gaussian filter, with width controlled by $\Sigma$.
A common choice for edge-preserving filtering is to choose color and position features $F^i=(x,y,r,g,b)^{\top}$.
This results in image smoothing without blurring across the edges.

The filter values $W^{i,j}$ change for every pixel pairs $i,j$ and depend on the image/video content.
And since the number of image/video pixels is usually large,
a naive implementation of
Eq.~\ref{eq:WtimesV} is prohibitive.
Due to the importance of this filtering operation, several fast algorithms
~\cite{adams2010fast,adams2009gaussian,paris2006fast,gastal2011domain} have been proposed,
that directly computes Eq.~\ref{eq:WtimesV} without explicitly building $W$ matrix.
One natural view that inspired several implementations was offered by~\cite{paris2006fast}, who viewed the bilateral filtering operation as a computation in a higher dimensional space.
Their observation was that bilateral filtering can be implemented by 1.~projecting $\mathbf{v}$ into a high-dimensional grid (\emph{splatting}) defined by features $F$, 2.~high-dimensional filtering (\emph{convolving}) the projected signal and 3.~projecting down the result at the points of interest (\emph{slicing}). The high-dimensional grid is also called \emph{bilateral space/grid}.
All these operations are linear and written as:

\vspace{-0.3cm}
\begin{equation}
\hat{\mathbf{v}} = S_{slice} B S_{splat} \mathbf{v},
\vspace{-0.1cm}
\end{equation}

where, $S_{splat}$ and $S_{slice}$ denotes the mapping to-from image pixels and bilateral grid, and $B$ denotes
convolution (traditionally Gaussian) in the bilateral space. The bilateral space has same dimensionality $g$ as
features $F^i$. The problem with this approach is that a standard $g$-dimensional convolution on a
regular grid requires handling of an exponential number of grid points.
This was circumvented by a special data structure, the permutohedral lattice as proposed in~\cite{adams2010fast}.
Effectively permutohedral filtering scales linearly with dimension, resulting in fast execution
time.

\begin{figure*}[th!]
\begin{center}
  \centerline{\includegraphics[width=0.8\textwidth]{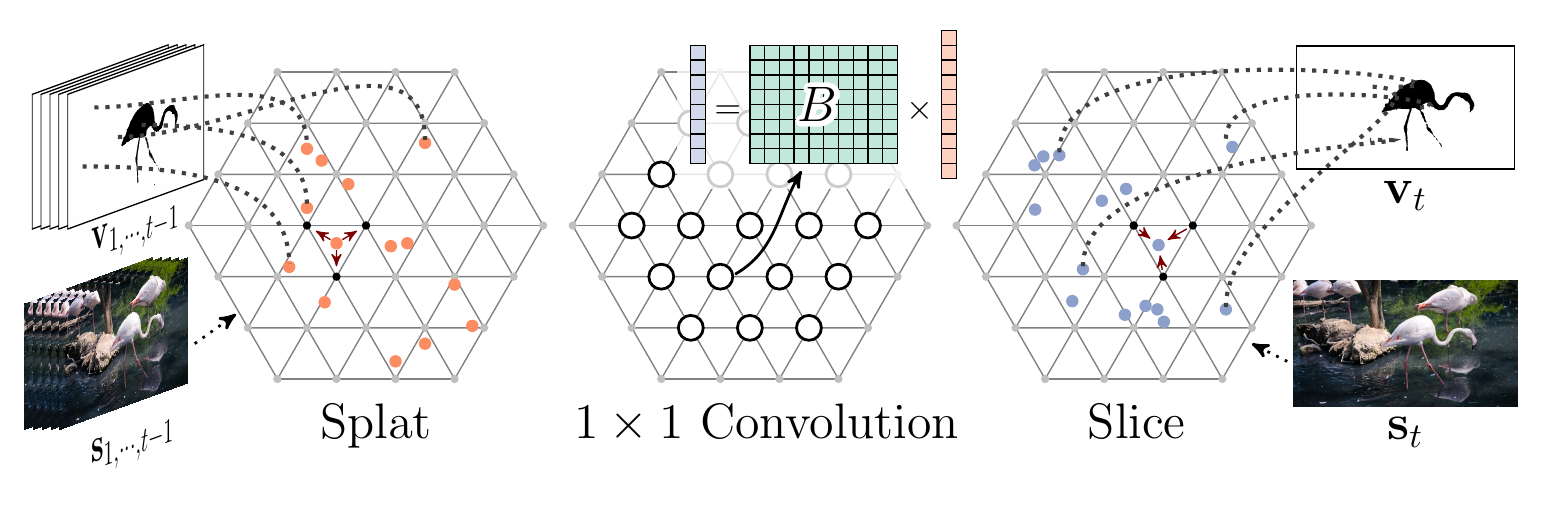}}
    \mycaption{Schematic of Fast Bilateral Filtering for Video Processing}
    {Mask probabilities from previous frames $\mathbf{v}_{1,\ldots,t-1}$ are splatted on to the
    lattice positions defined by the image features $F_1,F_2,\ldots,F_{t-1}$.
    The splatted result is convolved with a $1 \times 1$ filter $B$, and the filtered
    result is sliced back to the original image space to get $\mathbf{v}_t$ for the present frame.
    Input and output need not be $\mathbf{v}_t$, but can also be any intermediate neural network representation.
    $B$ is learned via back-propagation through these operations.}
    \label{fig:filter_illustration}
    \vspace{-1.0cm}
\end{center}
\end{figure*}

The recent work of~\cite{kiefel2014permutohedral,jampani_2016_cvpr} then generalized the bilateral filter in the
permutohedral lattice and demonstrated how it can be learned via back-propagation.
This allowed the construction of image-adaptive filtering operations into deep learning architectures, which we
will build upon. See Fig.~\ref{fig:filter_illustration} for a illustration of
2D permutohedral lattices. Refer to~\cite{adams2010fast} for more details on bilateral filtering
using permutohedral lattice and refer to~\cite{jampani_2016_cvpr} for details on
learning general permutohedral filters via back-propagation.

\section{Video Propagation Networks}\label{sec:vpn}

We aim to adapt the bilateral filtering operation to predict information forward in time, across video frames.
Formally, we work on a sequence of $h$ (color or grayscale) images $S = (\mathbf{s}_1, \mathbf{s}_2, \ldots, \mathbf{s}_h)$ and denote with $V = (\mathbf{v}_1, \mathbf{v}_2, \ldots, \mathbf{v}_h)$ a sequence of outputs, one per frame.
Consider as an example a sequence $\mathbf{v}_1,\ldots,\mathbf{v}_h$ of foreground masks for a moving object in the scene.
Our goal is to develop an online propagation method that can
predict $\mathbf{v}_t$, having observed the video up to frame $t$ and possibly previous $\mathbf{v}_{1,\ldots,t-1}$

\vspace{-0.3cm}
\begin{equation}
\mathcal{F}(\mathbf{v}_{t-1}, \mathbf{v}_{t-2}, \ldots; \mathbf{s}_t, \mathbf{s}_{t-1}, \mathbf{s}_{t-2},\ldots) = \mathbf{v}_t.
\vspace{-0.1cm}
\end{equation}

If training examples $\{(S_i,V_i) | i = 1,\ldots,l \}$ with full or partial knowledge of $\mathbf{v}$ are available, it is possible to learn $\mathcal{F}$ and for a complex and unknown input-output relationship, a deep CNN is a natural design choice. However, any learning based method has to face the challenge: the scene/camera motion and its
effect on $\mathbf{v}$. Since no motion in two different videos is the same, fixed-size static receptive fields of CNN
are insufficient. We propose to resolve this with video-adaptive filtering component, an adaption of the bilateral
filtering to videos.
Our Bilateral Network (Section~\ref{sec:bilateralnetwork}) has a connectivity that adapts to video sequences, its output is then fed into a spatial Network (Section~\ref{sec:spatialcnn}) that further refines the desired output.
The combined network layout of this VPN is depicted in Fig.~\ref{fig:net_illustration}.
It is a sequence of learnable bilateral and spatial filters that is efficient, trainable end-to-end and adaptive to the video input.

\subsection{Bilateral Network (BNN)}\label{sec:bilateralnetwork}
Several properties of bilateral filtering make it a perfect candidate for information propagation in videos.
In particular, our method is inspired by two main ideas that we extend in this work: joint bilateral upsampling~\cite{kopf2007joint} and learnable bilateral filters~\cite{jampani_2016_cvpr}. Although,
bilateral filtering has been used for filtering video data before~\cite{paris2008edge},
its use has been limited to fixed filter weights (say, Gaussian).

{\bf Fast Bilateral Upsampling across Frames} The idea of joint bilateral upsampling~\cite{kopf2007joint}
is to view upsampling as a filtering operation.
A high resolution guidance image is used to upsample a low-resolution result.
In short, a smaller number of input points are given $\{\mathbf{v}_{in}^i,F_{in}^i| i=1,\ldots,n_{in}\}$, for example a
segmentation result $\mathbf{v}_{in}$ at a lower resolution with the corresponding guidance image
features $F_{in}$. This is then scaled to a larger number of output points $\mathbf{v}_{out}$ with features $\{F_{out}^j|j=1,\ldots,n_{out}\}$
using the bilateral filtering operation, that is to compute Eq.~\ref{eq:WtimesV}, where the sum runs over all $n_{in}$
points and the output is computed for all $n_{out}$ positions ($W \in \mathbb{R}^{n_{in} \times n_{out}}$).

We will use this idea to propagate content from previous frames
($\mathbf{v}_{in} = \mathbf{v}_{1,\ldots,t-1}$) to the current frame ($\mathbf{v}_{out} = \mathbf{v}_t$).
The summation in Eq.~\ref{eq:WtimesV} now runs over \emph{all} previous frames and pixels.
This is illustrated in Fig.~\ref{fig:filter_illustration}. We take all previous frame results
$\mathbf{v}_{1,\ldots,t-1}$ and splat them into a lattice using the features $F_{1,\ldots,t-1}$ computed on video frames
$\mathbf{s}_{1,\ldots,t-1}$.
A filtering (described below) is then applied to every lattice point and the result is then sliced back using the
features $F_t$ of the current frame $\mathbf{s}_t$.
This result need not be the final $\mathbf{v}_t$, in fact we compute a filter bank of responses and continue
with further processing as will be discussed.

Standard bilateral features $F^i=(x,y,r,g,b)^{\top}$ used for images need not be optimal for videos.
A recent work of~\cite{kundu2016feature} propose to optimize bilateral feature spaces for
videos.
Instead, we choose to simply add frame index $t$ as an additional time feature yielding a 6 dimensional feature
vector $F^i=(x,y,r,g,b,t)^{\top}$ for every video pixel.
Imagine a video where an object moves to reveal some background.
Pixels of the object and background will be close spatially $(x,y)^\top$ and temporally $(t)$ but likely be of different color $(r,g,b)^\top$.
Therefore they will have no strong influence on each other (being splatted to distant positions in the six-dimensional bilateral space).
One can understand the filter to be adaptive to color changes across frames, only pixels that are static and have similar color have a strong influence on each other (end up nearby in the bilateral space).
In all our experiments, we used time $t$ as additional feature for information propagation across frames.

In addition to adding time $t$ as additional feature, we also experimented with using optical flow.
We make use of optical flow estimates (of the previous frames with respect to the current frame)
by warping pixel position features $(x,y)^\top$ of previous frames by their optical flow displacement vectors
$(u_x,u_y)^\top$ to $(x+u_x,y+u_y)^\top$.
If the perfect flow was available, the video frames could be warped into a common frame of reference. This would resolve the corresponding problem and make information propagation much easier.
We refer to the VPN model that uses modified positional features $(x+u_x,y+u_y)^\top$ as \emph{VPN-Flow}.

Another property of permutohedral filtering that we exploit
is that the \emph{input points need not lie on a regular grid} since the filtering
is done in the high-dimensional lattice. Instead of splatting millions of pixels on to the
lattice, we randomly sample or use superpixels and perform filtering using these sampled
points as input to the filter. In practice, we observe that this results in big computational
gains with minor drop in performance (more in Section~\ref{sec:videoseg}).

\begin{figure*}[th!]
\begin{center}
\centerline{\includegraphics[width=\textwidth]{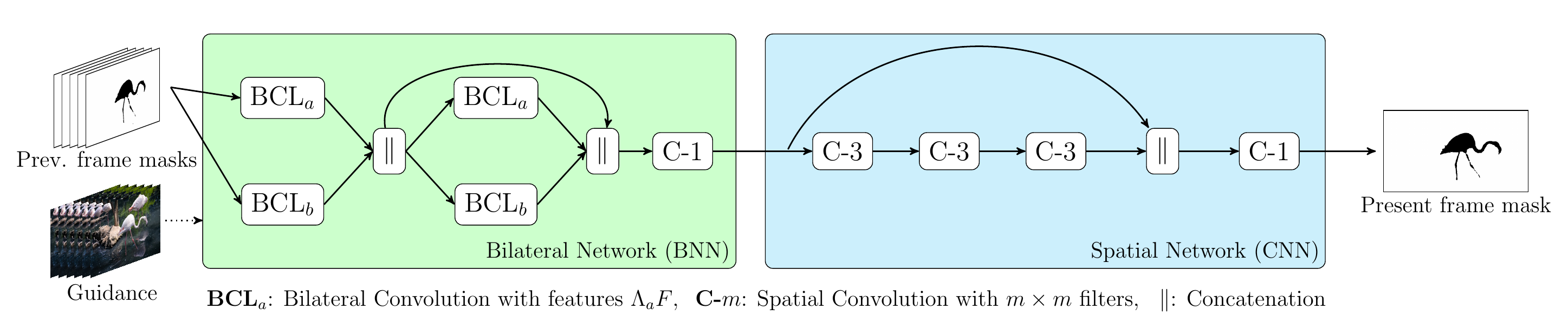}}
  \mycaption{Computation Flow of Video Propagation Network} {Bilateral networks (BNN) consist of a series of
  bilateral filterings interleaved with ReLU non-linearities. The filtered information from BNN is then passed
  into a
  spatial network (CNN) which refines the features with convolution layers interleaved with ReLU
  non-linearities, resulting in the prediction for the current frame.}
  \label{fig:net_illustration}
  \vspace{-1.0cm}
\end{center}
\end{figure*}

{\bf Learnable Bilateral Filters} Bilateral filters help in video-adaptive information propagation across
frames.
But the standard Gaussian filter may be insufficient and further, we would like to increase the capacity by using a filter bank instead of a single fixed filter.
We propose to use the technique of~\cite{jampani_2016_cvpr} to learn a filter bank in the permutohedral lattice using back-propagation.

The process works as follows.
A input video is used to determine the positions in the bilateral space to splat the input points
$\mathbf{v}^i \in \mathbf{v}_{1,\ldots,t-1}$ of the previous frames. In a general case, $\mathbf{v}^i$ need not be a scalar
and let us assume $\mathbf{v}^i \in \mathbb{R}^d$. The features $F_{1,\ldots,t}$ (e.g. $(x,y,r,g,b,t)^\top$) define the splatting matrix $S_{splat}$.~This leads to a number of vectors $\mathbf{v}_{splatted} = S_{splat}\mathbf{v}$, that lie on the permutohedral lattice, with dimensionality $\mathbf{v}^i_{splatted}\in\mathbb{R}^d$.
In effect, the splatting operation groups points that are close together, that is, they have similar $F^i,F^j$.
All lattice points are now filtered using a filter bank $B\in\mathbb{R}^{k\times d}$ which results in $k$ dimensional vectors on the lattice points.
These are sliced back to the $n_{out}$ points of interest (present video frame).
The values of $B$ are learned by back-propagation.
General parametrization of $B$ from~\cite{jampani_2016_cvpr,kiefel2014permutohedral} allows to have
any neighborhood size for the filters. Since constructing the neighborhood structure in
high-dimensions is time consuming, we choose to use $1 \times 1$ filters for speed reasons.
These three steps of \emph{splatting}, \emph{convolving} and \emph{slicing}
makes up one \emph{Bilateral Convolution Layer (BCL)} which we will stack and concatenate to form a Bilateral Network.
See Fig.~\ref{fig:filter_illustration} for a BCL illustration.

{\bf BNN Architecture} The Bilateral Network (BNN) is illustrated in the green box of
Fig.~\ref{fig:net_illustration}.
The input is a video sequence $S$ and the corresponding predictions $V$ up to frame
$t$. Those are filtered using two BCLs (BCL$_a$, BCL$_b$) with $32$ filters each.
For both BCLs, we use the same features $F^i$ but scale them with different diagonal matrices:
$\Lambda_a F^i,\Lambda_b F^i$. The feature scales ($\Lambda_a,\Lambda_b$) are found by validation.
The two $32$ dimensional outputs are concatenated, passed through a ReLU non-linearity and passed to a
second layer of two separate BCL filters that uses same feature spaces $\Lambda_a F^i,\Lambda_b F^i$.
The output of the second filter bank is then reduced using a $1\times 1$ spatial filter to map to
the original dimension $d$ of $\mathbf{v}$.
We investigated scaling frame inputs with an exponential time decay and found that, when processing
frame $t$, a re-weighting with $(\alpha \mathbf{v}_{t-1}, \alpha^2 \mathbf{v}_{t-2}, \alpha^3 \mathbf{v}_{t-3} \ldots)$ with
$0\le\alpha\le 1$ improved the performance a little bit.

In the experiments, we also included a simple BNN variant,
where no filters are applied inside the permutohedral space, just splatting and slicing
with the two layers BCL$_a$ and BCL$_b$ and adding the results.
We will refer to this model as \emph{BNN-Identity} as this is equivalent to using filter
$B$ that is identity matrix. It corresponds to an image adaptive smoothing of the inputs $V$.
We found this filtering to already have a positive effect in our experiments.

\vspace{-0.1cm}
\subsection{Spatial Network}\label{sec:spatialcnn}

The BNN was designed to propagate information from the previous frames to the present one,
respecting the scene and object motion.
We then add a small spatial CNN with 3 layers, each with $32$ filters of size $3\times 3$,
interleaved with ReLU non-linearities.
The final result is then mapped to the desired output of $\mathbf{v}_t$ using a $1\times 1$
convolution.
The main role of this spatial CNN is to refine the information in frame $t$.
Depending on the problem and the size of the available training data, other network designs are
conceivable. We use the same network architecture shown in Fig.~\ref{fig:net_illustration}
for all the experiments to demonstrate the generality of VPNs.

\vspace{-0.1cm}
\section{Experiments}
\label{sec:exps}

We evaluated VPN on three different propagation tasks: propagation of foreground masks, semantic
labels and color in videos. Our implementation runs in Caffe~\cite{jia2014caffe} using standard settings. We used Adam~\cite{kingma2014adam} stochastic optimization for training VPNs, multinomial-logistic loss for label propagation networks and Euclidean loss for training
color propagation networks. We use a fixed learning rate of 0.001 and choose the trained models
with minimum validation loss.
Runtime computations were performed using a
Nvidia TitanX GPU and a 6 core Intel i7-5820K CPU clocked at 3.30GHz machine.
The code is available online at http://varunjampani.github.io/vpn/.

\subsection{Video Object Segmentation}
\label{sec:videoseg}
We focus on the semi-supervised task of propagating
a given first frame foreground mask to all the video frames.
Object segmentation in videos is useful for several high level tasks such
as video editing, rotoscoping etc.

\vspace{-0.4cm}
\paragraph{Dataset} We use the recently published DAVIS dataset~\cite{Perazzi2016}
for experiments on this task.
It consists of 50 high-quality videos.
All the frames come with
high-quality per-pixel annotation of the foreground object.
For robust evaluation and to get results on all the dataset videos,
we evaluate our technique using 5-fold cross-validation.
We randomly divided the data into
5 folds, where in each fold, we used 35 videos for training, 5 for validation and
the remaining 10 for the testing. For the evaluation, we used the 3 metrics that
are proposed in~\cite{Perazzi2016}: Intersection over Union (IoU) score, Contour
accuracy ($\mathcal{F}$) score and temporal instability ($\mathcal{T}$) score. The widely
used IoU score is defined as $TP/(TP+FN+FP)$, where TP: True Positives; FN: False Negatives
and FP: False Positives. Refer to~\cite{Perazzi2016} for the definition of the other two metrics.

\begin{table}[t]
    \scriptsize
    \centering
    \begin{tabular}{p{1.5cm}>{\centering\arraybackslash}p{0.6cm}>{\centering\arraybackslash}
      p{0.6cm}>{\centering\arraybackslash}p{0.6cm}>{\centering\arraybackslash}p{0.6cm}>{\centering\arraybackslash}p{0.6cm}
      >{\centering\arraybackslash}p{0.6cm}}
        \toprule
        \scriptsize
        & F-1 & F-2 & F-3 & F-4 & F-5 & All\\ [0.1cm]
        \midrule
        BNN-Identity & 56.4 & 74.0 & 66.1 & 72.2 & 66.5 & 67.0 \\
        VPN-Stage1 & 58.2 & 77.7 & 70.4 & 76.0 & 68.1 & 70.1 \\
        VPN-Stage2 & \textbf{60.9} & \textbf{78.7} & \textbf{71.4} & \textbf{76.8} & \textbf{69.0} & \textbf{71.3} \\
        \bottomrule
        \\
    \end{tabular}
    \mycaption{5-Fold Validation on DAVIS Video Segmentation Dataset}
    {Average IoU scores for different models on the 5 folds.}
    \label{tbl:davis-folds}
\end{table}

\begin{table}[t]
    \scriptsize
    \centering
    \begin{tabular}{p{2.6cm}>{\centering\arraybackslash}p{0.6cm}>{\centering\arraybackslash}
      p{0.6cm}>{\centering\arraybackslash}p{0.6cm}>{\centering\arraybackslash}p{1.2cm}}
        \toprule
        \scriptsize
        & \textit{IoU$\uparrow$} & $\mathcal{F}\uparrow$ & $\mathcal{T}\downarrow$ & \textit{Runtime}(s) \\ [0.1cm]
        \midrule
        BNN-Identity & 67.0 & 67.1 & 36.3 & 0.21\\
        VPN-Stage1 & 70.1 & 68.4 & 30.1 & 0.48\\
        VPN-Stage2 & 71.3 & 68.9 & 30.2 & 0.75\\
        \midrule
        \multicolumn{4}{l}{\emph{With pre-trained models}} & \\
        DeepLab & 57.0 & 49.9 & 47.8 & 0.15 \\
        VPN-DeepLab & \textbf{75.0} & \textbf{72.4} & 29.5 & 0.63 \\
        \midrule
        OFL~\cite{tsaivideo} & 71.1 & 67.9 & 22.1 & $>$60\\
        BVS~\cite{marki2016bilateral} & 66.5 & 65.6 & 31.6 &  0.37\\
        NLC~\cite{faktor2014video} & 64.1 & 59.3 & 35.6 & 20\\
        FCP~\cite{perazzi2015fully} & 63.1 & 54.6 & 28.5 & 12\\
        JMP~\cite{fan2015jumpcut} & 60.7 & 58.6 & \textbf{13.2} & 12\\
        HVS~\cite{grundmann2010efficient} & 59.6 & 57.6 & 29.7 & 5\\
        SEA~\cite{ramakanth2014seamseg} & 55.6 & 53.3 & 13.7 & 6\\
        \bottomrule
        \\
    \end{tabular}
    \mycaption{Results of Video Object Segmentation on DAVIS dataset}
    {Average IoU score, contour accuracy ($\mathcal{F}$),
    temporal instability ($\mathcal{T}$) scores, and average runtimes (in seconds)
    per frame for different VPN models along with recent published
    techniques for this task. VPN runtimes also include superpixel computation (10ms).
    Runtimes of other methods are taken from~\cite{marki2016bilateral,perazzi2015fully,tsaivideo}
    which are indicative and are not directly comparable to our runtimes.
    Runtime of VPN-Stage2 includes the runtime of VPN-Stage1 which in turn includes the runtime of BNN-Identity.
    Runtime of VPN-DeepLab model includes the runtime of DeepLab.}
    \label{tbl:davis-main}
    \vspace{-0.3cm}
\end{table}

\vspace{-0.4cm}
\paragraph{VPN and Results} In this task, we only have access to foreground mask
for the first frame $\mathbf{v}_1$.
For the ease of training VPN, we obtain initial set of predictions with
\emph{BNN-Identity}. We sequentially apply \emph{BNN-Identity} at each frame
and obtain an initial set of foreground masks for the entire video.
These BNN-Identity propagated masks are then used as inputs to train a VPN to
predict the refined masks at each frame. We refer to this
VPN model as \emph{VPN-Stage1}. Once VPN-Stage1 is trained, its refined
mask predictions are in-turn used as inputs to train another VPN model which we
refer to as \emph{VPN-Stage2}. This resulted in further refinement of foreground
masks. Training further stages did not result in any improvements. Instead, one
could consider VPN as a RNN unit processing one frame after another. But, due to
GPU memory constraints, we opted for stage-wise training.

\begin{figure}[t!]
\begin{center}
  \centerline{\includegraphics[width=0.6\columnwidth]{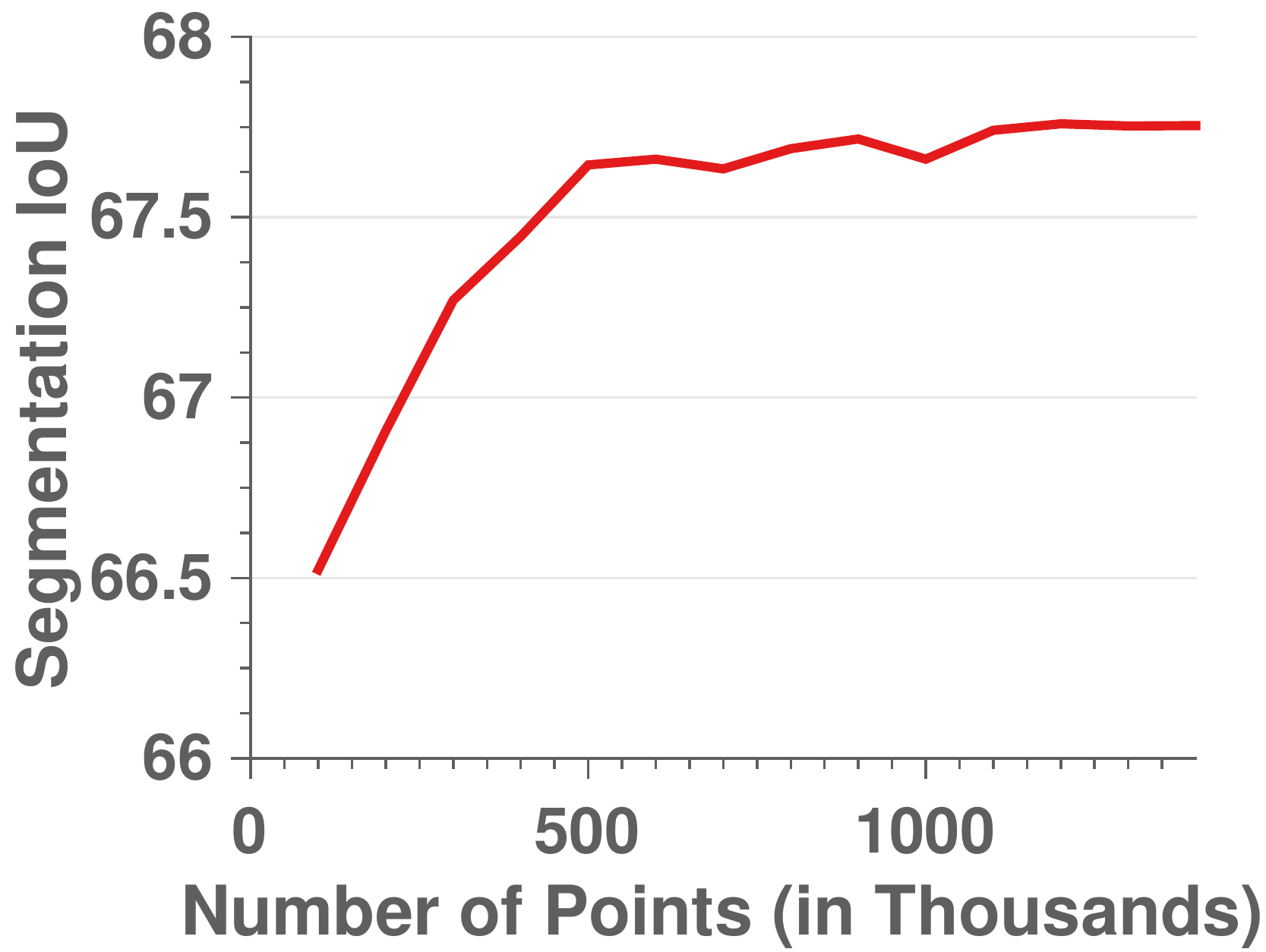}}
    \mycaption{Random Sampling of Input Points vs. IoU}
    {The effect of randomly sampling points from input video frames on object
    segmentation IoU of BNN-Identity on DAVIS dataset.
    The points sampled are out of $\approx$2 million points from the previous 5 frames.}
    \label{fig:acc_vs_points}
\end{center}
\vspace{-1.2cm}
\end{figure}

Following the recent work of~\cite{marki2016bilateral} on video object segmentation,
we used $F^i=(x,y,Y,Cb,Cr,t)^\top$ features with YCbCr color features for bilateral filtering.
To be comparable with one of the fastest state-of-the-art technique~\cite{marki2016bilateral},
we do \emph{not} use any optical flow information. First, we analyze the performance of BNN-Identity by changing the number of randomly sampled input points. Figure~\ref{fig:acc_vs_points} shows how the segmentation IoU changes with
the number of sampled points (out of 2 million points) from the previous frames.
The IoU levels out after sampling 25\% of the points. For
further computational efficiency, we used superpixel sampling instead of random
sampling. Compared to random sampling,
usage of superpixels reduced the IoU slightly (0.5), while reducing the
number of input points by a factor of 10.
We used 12000 SLIC~\cite{achanta2012slic} superpixels from each frame
computed using the fast GPU implementation from~\cite{gSLICr_2015}. As an input to VPN,
we use the mask probabilities of previous 9 frames as we observe
no improvements with more frames. We set $\alpha = 0.5$ and the
feature scales ($\Lambda_a,\Lambda_b$) are presented in Tab.~\ref{tbl:parameters_supp}.

Table~\ref{tbl:davis-folds} shows the IoU scores for each of the 5 folds and
Tab.~\ref{tbl:davis-main} shows the overall scores and runtimes of different VPN
models along with the best performing techniques.
The performance improved consistently across all 5 folds with the addition of new VPN stages.~BNN-Identity already performed reasonably well.
VPN outperformed the present fastest
BVS method~\cite{marki2016bilateral} by a significant margin on all
the performance measures while being comparable in runtime.
VPN perform marginally better than OFL method~\cite{tsaivideo}
while being at least 80$\times$ faster and OFL relies on optical flow whereas we
obtain similar performance without using any optical flow.
~Further, VPN has the advantage of doing online processing
as it looks only at previous frames whereas BVS processes entire video at once.

\begin{figure}[th!]
\begin{center}
  \centerline{\includegraphics[width=\columnwidth]{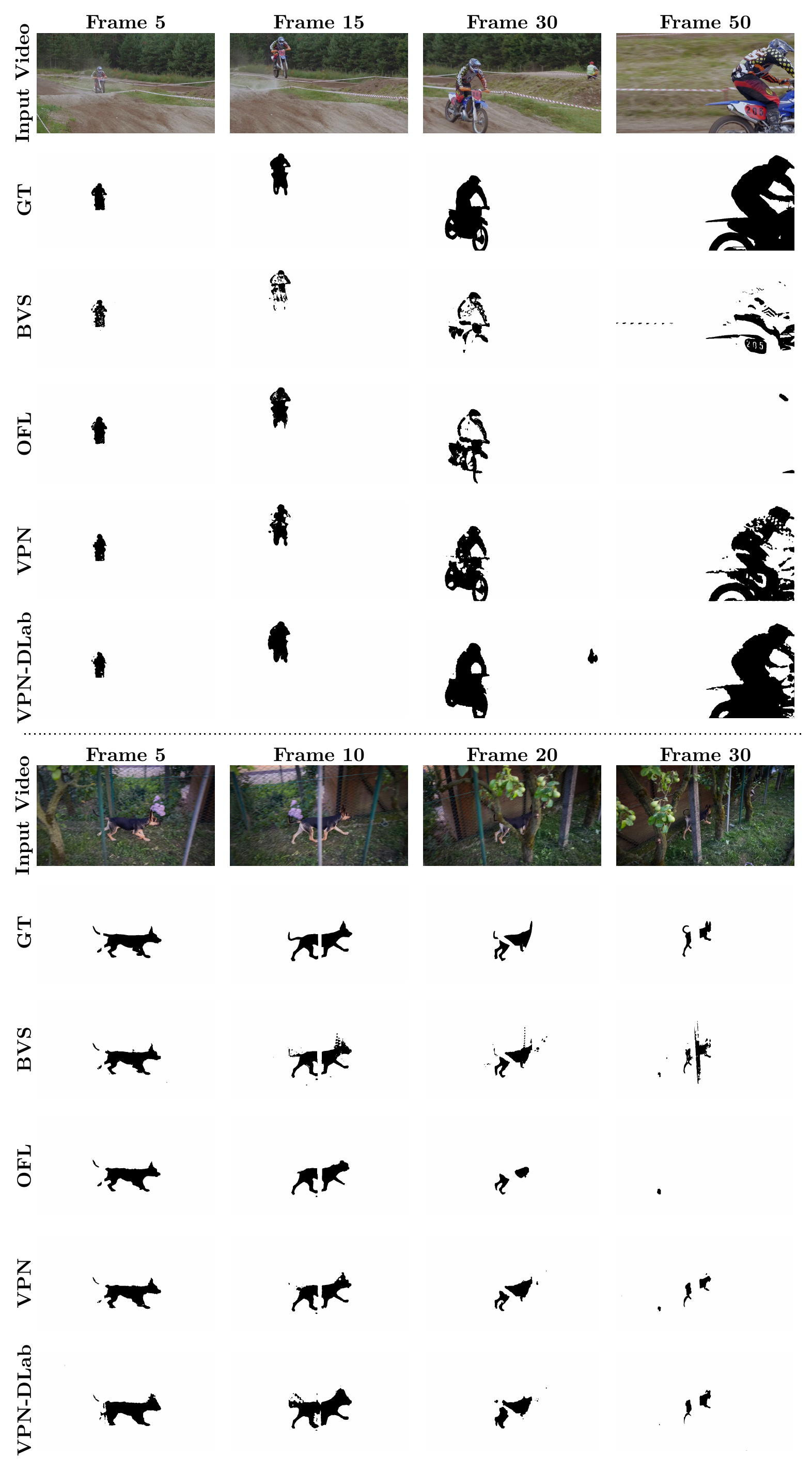}}
    \mycaption{Video Object Segmentation}
    {Shown are the different frames in example videos with the corresponding
    ground truth (GT) masks, predictions from BVS~\cite{marki2016bilateral},
    OFL~\cite{tsaivideo}, VPN (VPN-Stage2) and VPN-DLab (VPN-DeepLab) models.}
    \label{fig:video_seg_visuals}
\end{center}
\vspace{-1.0cm}
\end{figure}

\vspace{-0.5cm}
\paragraph{Augmentation of Pre-trained Models} One of the main advantages of VPN
is that it is end-to-end trainable and can be easily integrated into
other deep networks. To demonstrate this, we augmented VPN architecture
with standard DeepLab segmentation network~\cite{chen2014semantic}.
We replaced the last classification layer of DeepLab-LargeFOV model
to output 2 classes (foreground and background)
in our case and bi-linearly upsampled the resulting low-resolution probability map to
the original image dimension. 5-fold fine-tuning of the DeepLab model on DAVIS dataset
resulted in the average IoU of 57.0 and other scores are shown in Tab.~\ref{tbl:davis-main}.
To construct a joint model, the outputs from
the DeepLab and the bilateral network (in VPN) are concatenated and then passed on to the spatial CNN.
In other words, the bilateral network propagates label information from previous frames to the present
frame, whereas the DeepLab network does the prediction for the present frame. The results
of both are then combined and refined by the spatial network in the VPN.
We call this `VPN-DeepLab' model. We trained this model end-to-end and observed big
improvements in performance. As shown in Tab.~\ref{tbl:davis-main}, the VPN-DeepLab
model has the IoU score of 75.0 which is a significant improvement over the published results.
The total runtime of VPN-DeepLab is only 0.63s which makes this also one of
the fastest techniques. Figure~\ref{fig:video_seg_visuals} shows
some qualitative results with more in Figs.~\ref{fig:video_seg_small_supp},~\ref{fig:video_seg_pos_supp}
and~\ref{fig:video_seg_neg_supp}.
One can obtain better VPN performance with using better superpixels and
also incorporating optical flow, but this increases runtime as well.
Visual results indicate that learned VPN is able to retain foreground masks even
with large variations in viewpoint and object size.

\subsection{Semantic Video Segmentation}

This is the task of assigning semantic label to every video pixel.
Since the semantics between adjacent frames does not change
radically, intuitively, propagating semantics across frames should improve
the segmentation quality of each frame. Unlike video object segmentation,
where the mask for the first frame is given, we approach semantic video segmentation
in a fully automatic fashion. Specifically, we start
with the unary predictions of standard CNNs and use VPN for propagating semantics across the frames.

\vspace{-0.4cm}
\paragraph{Dataset} We use the CamVid dataset \cite{brostow2009semantic} that contains 4 high
quality videos captured at 30Hz while the semantically labeled 11-class ground truth is
provided at 1Hz. While the original dataset comes at a resolution of 960$\times$720,
we operate on a resolution of 640$\times$480 as in~\cite{yu2015multi,kundu2016feature}.
We use the same splits as in~\cite{sturgess2009combining} resulting in
367, 100 and 233 frames with ground truth for training, validation and testing.

\vspace{-0.4cm}
\paragraph{VPN and Results} Since we already have CNN predictions for every
frame, we train a VPN that takes the CNN predictions of previous \emph{and} present
frames as input and predicts the refined semantics for the present frame.
We compare with a state-of-the-art CRF approach~\cite{kundu2016feature}
which we refer to as FSO-CRF. We also experimented with
optical flow in VPN and refer that model as \emph{VPN-Flow}. We used the fast DIS
optical flow~\cite{kroeger2016fast} and modify the positional features of previous frames.
We used superpixels computed with Dollar et al.~\cite{dollariccv13edges} as
gSLICr~\cite{gSLICr_2015} has introduced artifacts.

\begin{table}[t]
    \scriptsize
    \centering
    \begin{tabular}{p{3.5cm}>{\centering\arraybackslash}p{1.2cm}>{\centering\arraybackslash}p{1.5cm}}
        \toprule
        \scriptsize
        & \textit{IoU} & \textit{Runtime}(s) \\ [0.1cm]
        \midrule
        CNN-1 from ~\cite{yu2015multi} & 65.3 & 0.38\\
        + FSO-CRF~\cite{kundu2016feature} & 66.1 & \textbf{$>$}10\\
        + BNN-Identity  & 65.3 & 0.31\\
        + BNN-Identity-Flow  & 65.5 & 0.33\\
        + VPN (Ours) & 66.5 & 0.35\\
        + VPN-Flow (Ours) & \textbf{66.7} & 0.37\\
        \midrule
        CNN-2 from ~\cite{richter2016playing} & 68.9 & 0.30\\
        + VPN-Flow (Ours) & \textbf{69.5} & 0.38\\
        \bottomrule
        \\
    \end{tabular}
    \mycaption{Results of Semantic Segmentation on the CamVid Dataset}{
    Average IoU and runtimes (in seconds)
    per frame of different models on \textit{test} split.
    Runtimes exclude CNN computations which are shown separately.
    VPN and BNN-Identity runtimes include superpixel computation
    of 0.23s (large portion of runtime).}
    \label{tbl:camvid}
    \vspace{-0.5cm}
\end{table}

\begin{figure}[th!]
  \vspace{-0.3cm}
\begin{center}
  \centerline{\includegraphics[width=\columnwidth]{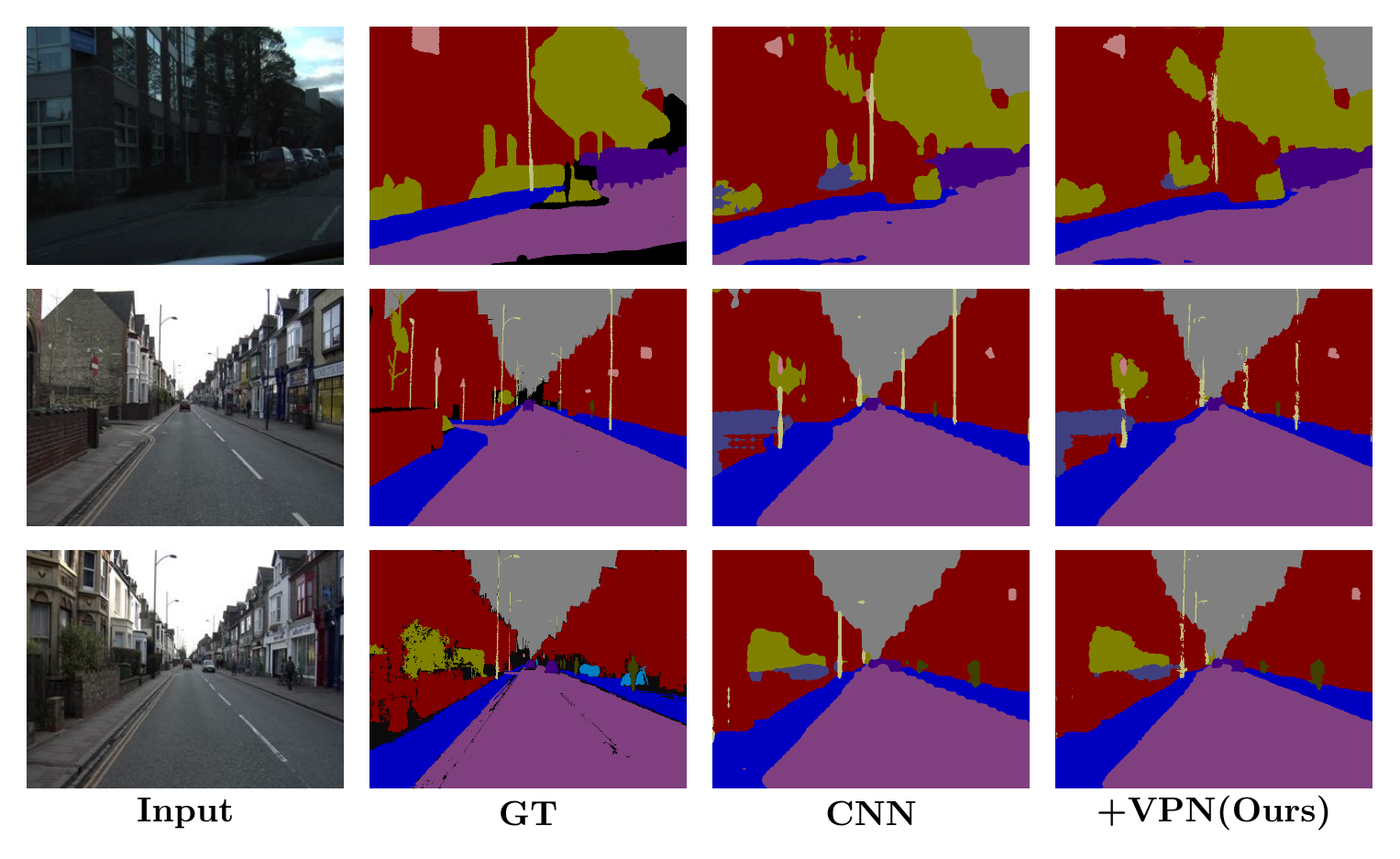}}
    \mycaption{Semantic Video Segmentation}
    {Input video frames and the corresponding ground truth (GT)
    segmentation together with the predictions of CNN~\cite{yu2015multi} and with
    VPN-Flow.}
    \label{fig:semantic_visuals}
\end{center}
\vspace{-0.7cm}
\end{figure}

We experimented with predictions from two different CNNs:
One is with dilated convolutions~\cite{yu2015multi} (CNN-1) and another one~\cite{richter2016playing} (CNN-2)
is trained with the additional video game data,
which is the present state-of-the-art on this dataset.
For CNN-1 and CNN-2, using 2 and 3 previous frames respectively as input
to VPN is found to be optimal. Other parameters of VPN are presented
in Tab.~\ref{tbl:parameters_supp}. Table~\ref{tbl:camvid} shows quantitative results.
Using BNN-Identity only slightly improved the performance whereas training the
entire VPN significantly improved the CNN-1 performance by over 1.2 IoU, with both
VPN and VPN-Flow. Moreover, VPN is at least 25$\times$ faster, and simpler to use
compared to the optimization based FSO-CRF which relies on
LDOF optical flow~\cite{brox2009large}, long-term tacks~\cite{sundaram2010dense} and
edges~\cite{dollar2015fast}.
Replacing bilateral filters with spatial filters in VPN improved the CNN-1 performance by only
0.3 IoU showing the importance of video-adaptive filtering.
We further improved the performance of the state-of-the-art CNN-2~\cite{richter2016playing}
with VPN-Flow model. Using better optical flow estimation
might give even better results. Figure~\ref{fig:semantic_visuals} shows some qualitative
results with more in Fig.~\ref{fig:semantic_visuals_supp}.

\vspace{-0.4cm}
\subsection{Video Color Propagation}

We also evaluate VPNs on a regression task of propagating color information in a grayscale video.
Given the color image for the first video frame, the task is to propagate the color to the entire
video.
For experiments on this task, we again used the DAVIS segmentation dataset~\cite{Perazzi2016} with the
first 25 frames from each video. We randomly divided the dataset into 30 train,
5 validation and 15 test videos.

\begin{figure}[th!]
\begin{center}
  \centerline{\includegraphics[width=\columnwidth]{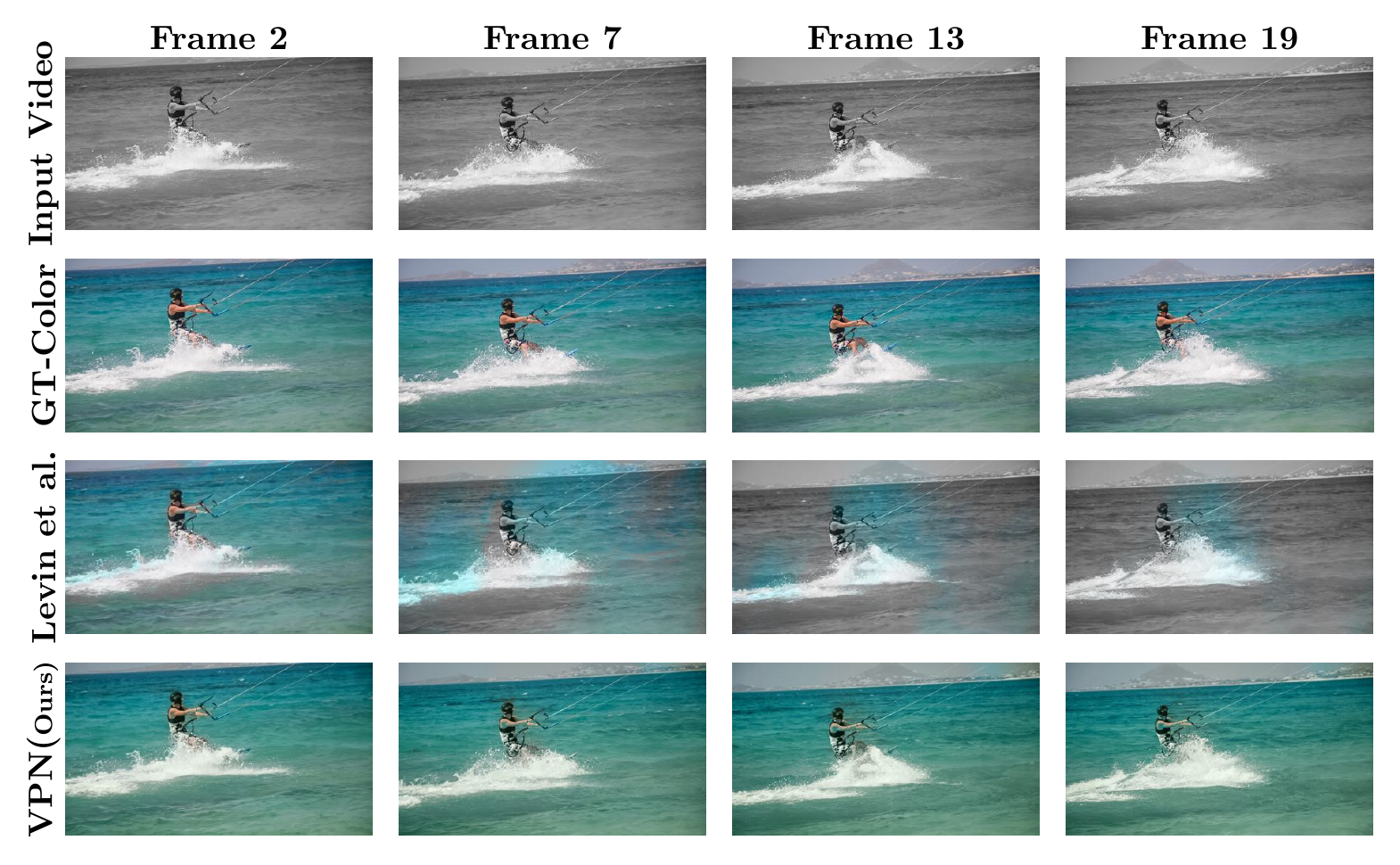}}
    \mycaption{Video Color Propagation}
    {Input grayscale video frames and corresponding ground-truth (GT) color images
    together with color predictions of Levin et al.~\cite{levin2004colorization} and VPN-Stage1 models.}
    \label{fig:color_visuals}
\end{center}
\vspace{-1.0cm}
\end{figure}

We work with YCbCr representation of images and propagate CbCr values from previous
frames with pixel intensity, position and time features as guidance for VPN.
The same strategy as in object segmentation is used, where an initial
set of color propagated results is obtained with BNN-Identity and then used to trained a VPN-Stage1 model.
Training further VPN stages did not improve the performance.
We use 300K randomly sampled points from previous 3 frames as input
to the VPN network. Table~\ref{tbl:color} shows the PSNR results.
We also show a baseline result of~\cite{levin2004colorization} that
does graph based optimization using optical flow. We used fast
DIS optical flow~\cite{kroeger2016fast} in the baseline method~\cite{levin2004colorization}
and we did not observe significant differences with using LDOF optical flow~\cite{brox2009large}.
Figure~\ref{fig:color_visuals} shows a visual result with more
in Fig.~\ref{fig:color_visuals_supp}.
VPN works reliably better than~\cite{levin2004colorization} while being 20$\times$ faster.
The method of~\cite{levin2004colorization} relies heavily on optical flow
and so the color drifts away with incorrect flow. We observe that our method also bleeds color
in some regions especially when there are large viewpoint changes.
We could not compare against recent color propagation techniques
~\cite{heu2009image,sheng2014video} as their codes are not available online.
This application shows general applicability of VPNs in propagating different
kinds of information.

\begin{table}[t]
    \scriptsize
    \centering
    \begin{tabular}{p{2.5cm}>{\centering\arraybackslash}p{1.2cm}>{\centering\arraybackslash}p{1.5cm}}
        \toprule
        \scriptsize
        & \textit{PSNR} & \textit{Runtime}(s) \\ [0.1cm]
        \midrule
        BNN-Identity & 27.89 & 0.29\\
        VPN-Stage1 & \textbf{28.15} & 0.90\\
        \midrule
        Levin et al.~\cite{levin2004colorization} & 27.11 & 19\\
        \bottomrule
        \\
    \end{tabular}
    \mycaption{Results of Video Color Propagation}{Average Peak Signal-to-Noise Ratio (PSNR) and runtimes of
    different methods for video color propagation on images from DAVIS dataset.}
    \label{tbl:color}
    \vspace{-0.6cm}
\end{table}

\vspace{-0.4cm}
\section{Conclusion}
\label{sec:conclusion}

We proposed a fast, scalable and generic neural network approach
for propagating information across video frames.~The VPN uses
bilateral network for long-range video-adaptive propagation of information from previous
frames to the present frame which is then refined by a spatial network.
Experiments on diverse tasks show that VPNs, despite being generic, outperformed
the current state-of-the-art task-specific methods. At the core of our technique
is the exploitation and modification of learnable bilateral filtering for the use
in video processing. We used a simple VPN architecture to showcase the generality.
Depending on the problem and the availability of data, using more filters or deeper layers
would result in better performance. In this work, we manually tuned the feature scales which
could be amendable to learning. Finding optimal yet fast-to-compute bilateral features for
videos together with the learning of their scales is an important future
research direction.

\vspace{-0.4cm}
{\small
\paragraph{Acknowledgments} We thank Vibhav Vineet for providing the trained
image segmentation CNN models for CamVid dataset.
}

\vspace{-0.3cm}
{\small
\bibliographystyle{ieee}
\bibliography{references}
}

\clearpage

\onecolumn

\appendix
\counterwithin{figure}{section}
\counterwithin{table}{section}

\section{Parameters and Additional Results}

\begin{figure}[th!]
\begin{center}
  \centerline{\includegraphics[width=\textwidth]{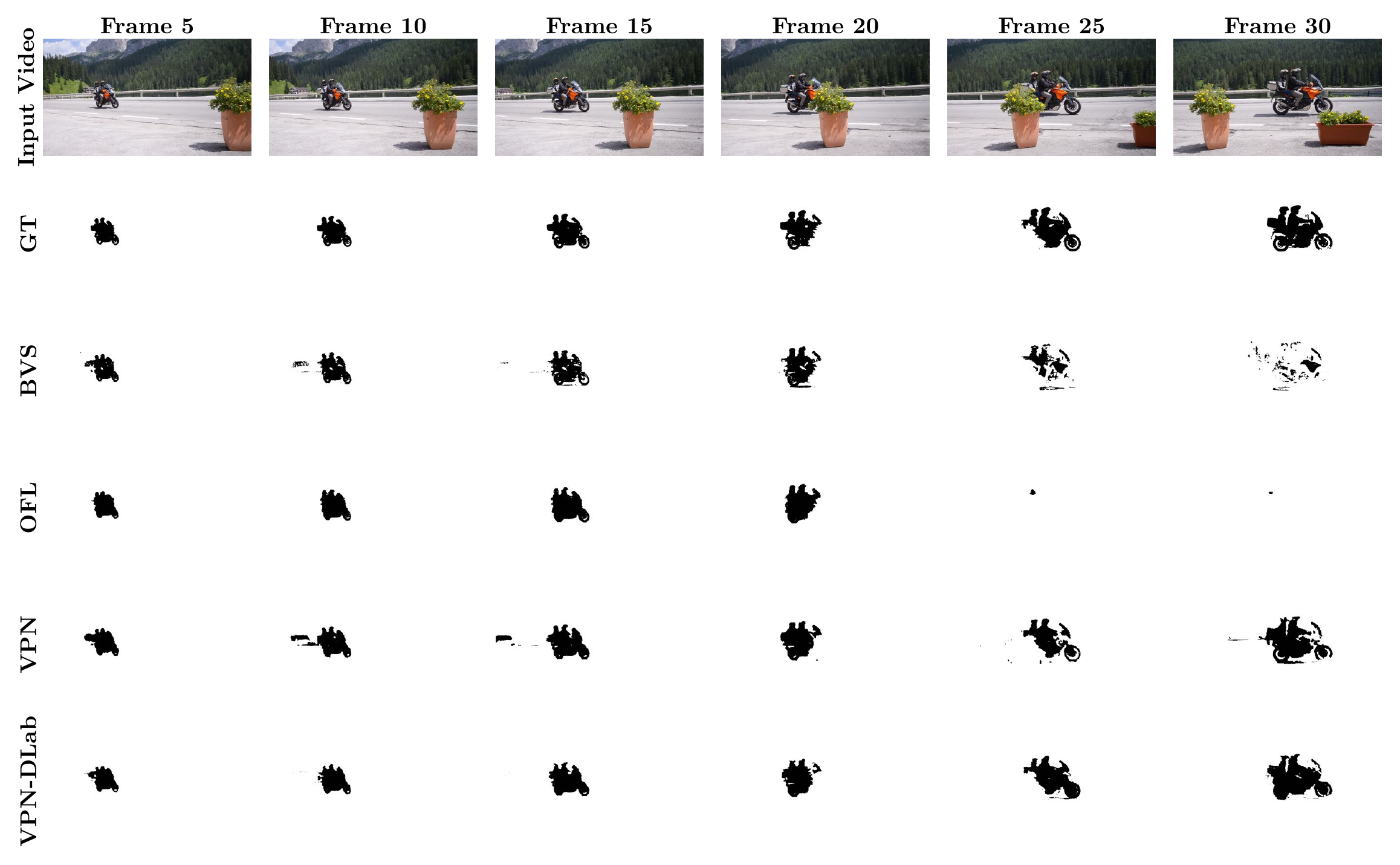}}
    \mycaption{Video Object Segmentation}
    {Shown are the different frames in example videos with the corresponding
    ground truth (GT) masks, predictions from BVS~\cite{marki2016bilateral},
    OFL~\cite{tsaivideo}, VPN (VPN-Stage2) and VPN-DLab (VPN-DeepLab) models.}
    \label{fig:video_seg_small_supp}
\end{center}
\vspace{-1.0cm}
\end{figure}

\begin{figure}[th!]
\begin{center}
  \centerline{\includegraphics[width=0.7\textwidth]{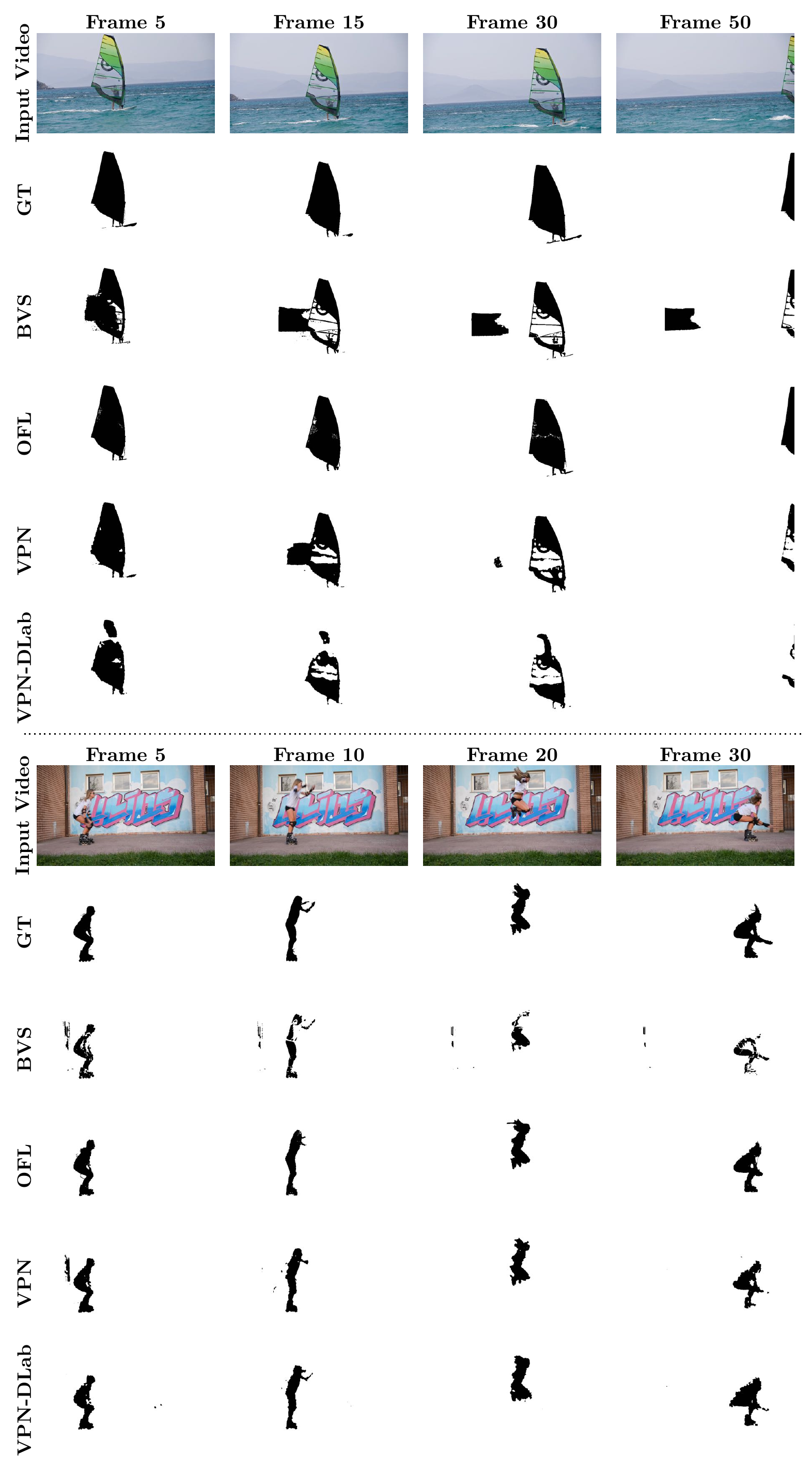}}
    \mycaption{Video Object Segmentation}
    {Shown are the different frames in example videos with the corresponding
    ground truth (GT) masks, predictions from BVS~\cite{marki2016bilateral},
    OFL~\cite{tsaivideo}, VPN (VPN-Stage2) and VPN-DLab (VPN-DeepLab) models.}
    \label{fig:video_seg_pos_supp}
\end{center}
\vspace{-1.0cm}
\end{figure}

\begin{figure}[th!]
\begin{center}
  \centerline{\includegraphics[width=0.7\textwidth]{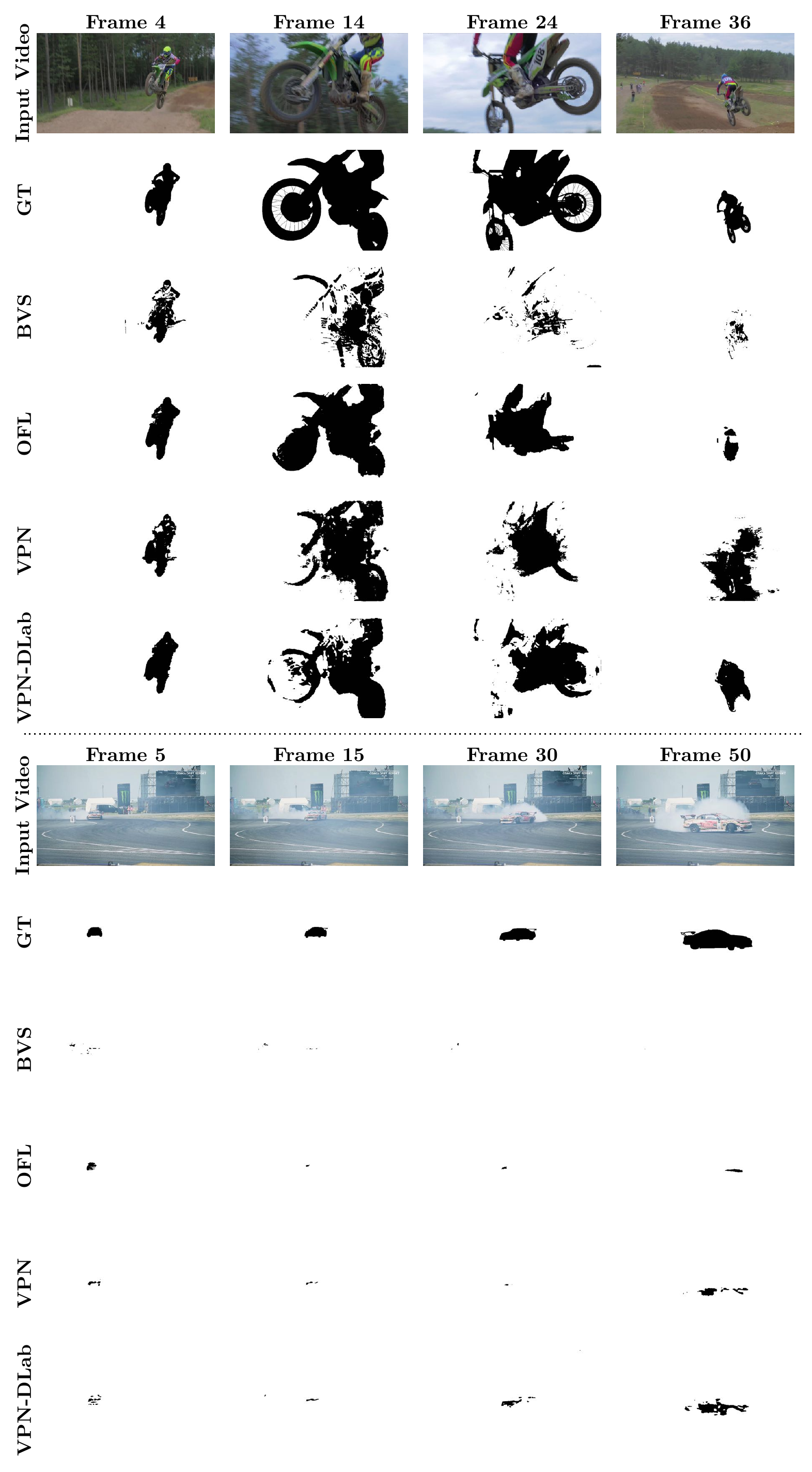}}
    \mycaption{Failure Cases for Video Object Segmentation}
    {Shown are the different frames in example videos with the corresponding
    ground truth (GT) masks, predictions from BVS~\cite{marki2016bilateral},
    OFL~\cite{tsaivideo}, VPN (VPN-Stage2) and VPN-DLab (VPN-DeepLab) models.}
    \label{fig:video_seg_neg_supp}
\end{center}
\vspace{-1.0cm}
\end{figure}

\begin{figure}[th!]
\begin{center}
  \centerline{\includegraphics[width=0.9\textwidth]{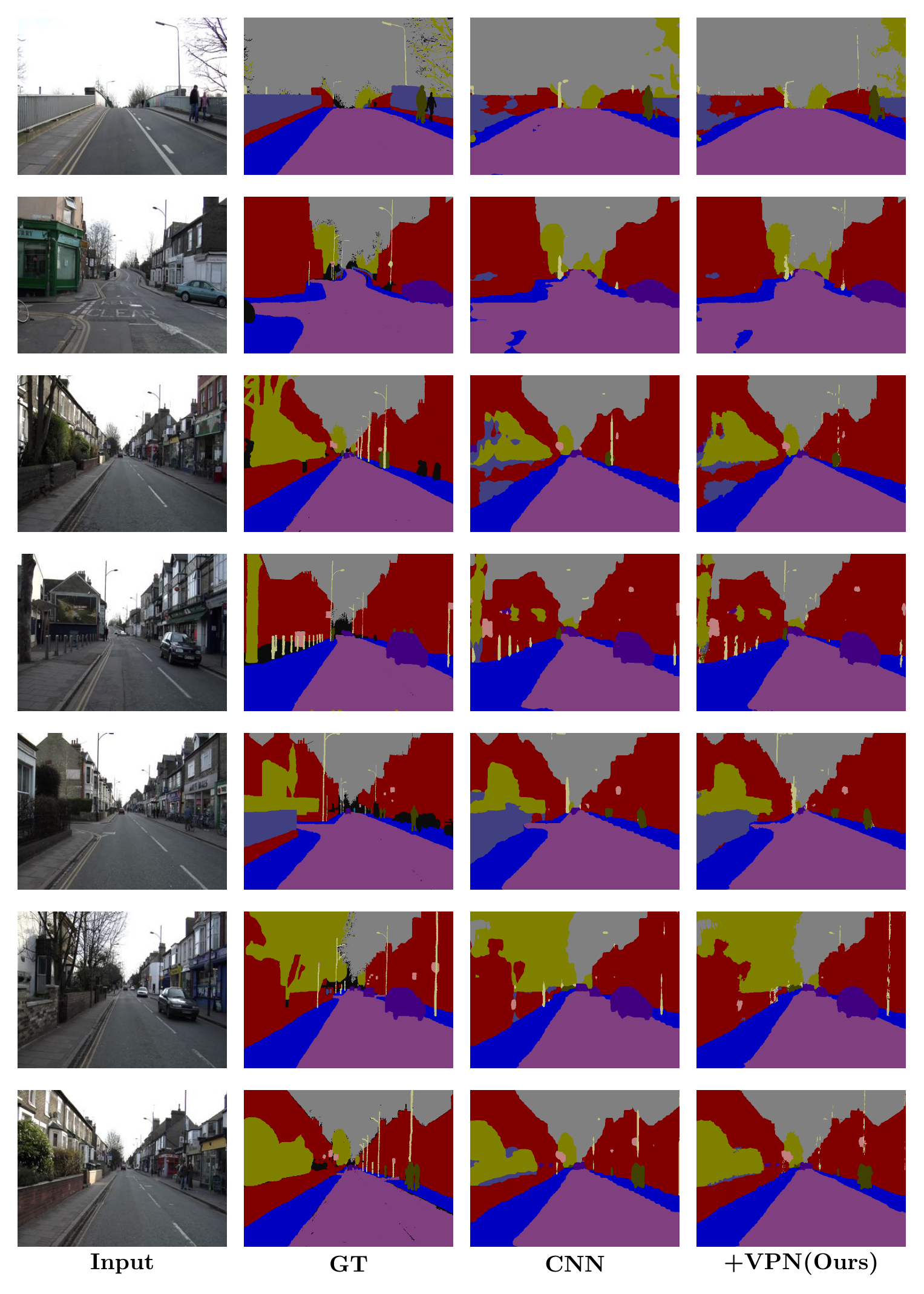}}
    \mycaption{Semantic Video Segmentation}
    {Input video frames and the corresponding ground truth (GT)
    segmentation together with the predictions of CNN~\cite{yu2015multi} and with
    VPN-Flow.}
    \label{fig:semantic_visuals_supp}
\end{center}
\vspace{-0.7cm}
\end{figure}

\begin{figure}[th!]
\begin{center}
  \centerline{\includegraphics[width=\textwidth]{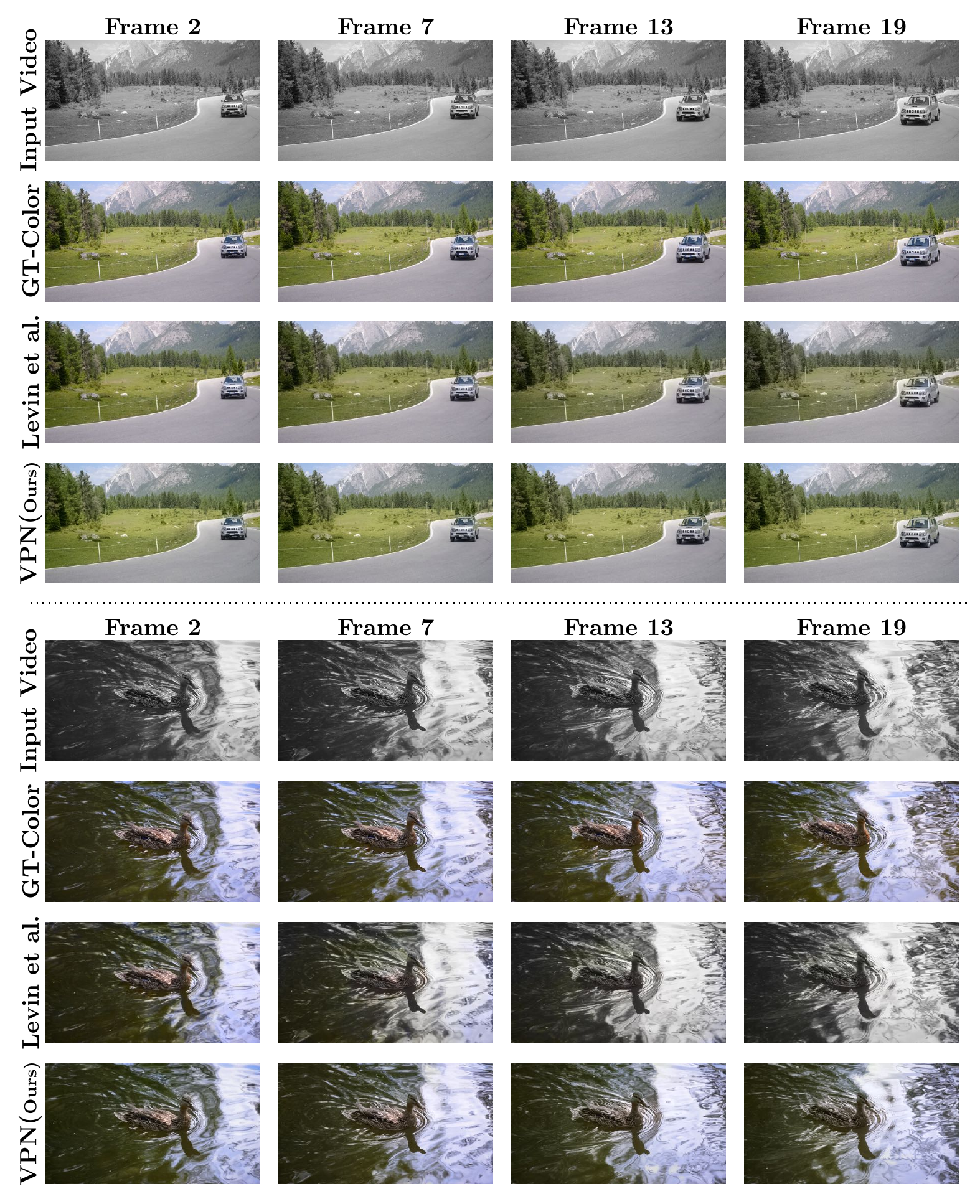}}
  \mycaption{Video Color Propagation}
  {Input grayscale video frames and corresponding ground-truth (GT) color images
  together with color predictions of Levin et al.~\cite{levin2004colorization} and VPN-Stage1 models.}
  \label{fig:color_visuals_supp}
\end{center}
\vspace{-0.7cm}
\end{figure}

In this appendix, we present experiment protocols and additional qualitative results for experiments
on video object segmentation, semantic video segmentation and video color
propagation. Table~\ref{tbl:parameters_supp} shows the feature scales and other parameters used in different experiments.
Figures~\ref{fig:video_seg_small_supp},~\ref{fig:video_seg_pos_supp} show some qualitative results on video object segmentation
with some failure cases in Fig.~\ref{fig:video_seg_neg_supp}.
Figure~\ref{fig:semantic_visuals_supp} shows some qualitative results on semantic video segmentation and
Fig.~\ref{fig:color_visuals_supp} shows results on video color propagation.

\newcolumntype{L}[1]{>{\raggedright\let\newline\\\arraybackslash\hspace{0pt}}b{#1}}
\newcolumntype{C}[1]{>{\centering\let\newline\\\arraybackslash\hspace{0pt}}b{#1}}
\newcolumntype{R}[1]{>{\raggedleft\let\newline\\\arraybackslash\hspace{0pt}}b{#1}}

\begin{table*}[h]
\scriptsize
  \centering
    \begin{tabular}{L{3.2cm} L{3.0cm} L{2.8cm} L{2.8cm} C{0.5cm} C{1.0cm} L{1.2cm}}
      \toprule
\textbf{Experiment} & \textbf{Feature Type} & \textbf{Feature Scale-1, $\Lambda_a$} & \textbf{Feature Scale-2, $\Lambda_b$} & \textbf{$\alpha$} & \textbf{Input Frames} & \textbf{Loss Type} \\
      \midrule
      \textbf{Video Object Segmentation} & ($x,y,Y,Cb,Cr,t$) & (0.02,0.02,0.07,0.4,0.4,0.01) & (0.03,0.03,0.09,0.5,0.5,0.2) & 0.5 & 9 & Logistic\\
      \midrule
      \textbf{Semantic Video Segmentation} & & & & & \\
      \textbf{with CNN1~\cite{yu2015multi}-NoFlow} & ($x,y,R,G,B,t$) & (0.08,0.08,0.2,0.2,0.2,0.04) & (0.11,0.11,0.2,0.2,0.2,0.04) & 0.5 & 3 & Logistic \\
      \textbf{with CNN1~\cite{yu2015multi}-Flow} & ($x+u_x,y+u_y,R,G,B,t$) & (0.11,0.11,0.14,0.14,0.14,0.03) & (0.08,0.08,0.12,0.12,0.12,0.01) & 0.65 & 3 & Logistic\\
      \textbf{with CNN2~\cite{richter2016playing}-Flow} & ($x+u_x,y+u_y,R,G,B,t$) & (0.08,0.08,0.2,0.2,0.2,0.04) & (0.09,0.09,0.25,0.25,0.25,0.03) & 0.5 & 4 & Logistic\\
      \midrule
      \textbf{Video Color Propagation} & ($x,y,I,t$)  & (0.04,0.04,0.2,0.04) & No second kernel & 1 & 4 & MSE\\
      \bottomrule
      \\
    \end{tabular}
    \mycaption{Experiment Protocols} {Experiment protocols for the different experiments presented in this work. \textbf{Feature Types}:
    Feature spaces used for the bilateral convolutions, with position ($x,y$) and color
    ($R,G,B$ or $Y,Cb,Cr$) features $\in [0,255]$. $u_x$, $u_y$ denotes optical flow with respect
    to the present frame and $I$ denotes grayscale intensity.
    \textbf{Feature Scales ($\Lambda_a, \Lambda_b$)}: Cross-validated scales for the features used.
    \textbf{$\alpha$}: Exponential time decay for the input frames.
    \textbf{Input Frames}: Number of input frames for VPN.
    \textbf{Loss Type}: Type
     of loss used for back-propagation. ``MSE'' corresponds to Euclidean mean squared error loss and ``Logistic'' corresponds to multinomial logistic loss.}
  \label{tbl:parameters_supp}
\end{table*}

\end{document}